\documentclass[10pt,journal,final]{IEEEtran}
\usepackage{times}
\usepackage{latexsym}
\usepackage{times}
\usepackage{float}
\usepackage{latexsym}
\usepackage{multirow}
\usepackage{url}
\usepackage{tabularx}
\usepackage{latexsym}
\usepackage{balance}
\usepackage{bm}
\usepackage{amssymb}
\usepackage{amsmath}
\usepackage{makecell}
\usepackage{multirow}
\usepackage{subfig}
\usepackage{graphicx} 
\usepackage{color}
\usepackage{colortbl}
\usepackage{textcomp}
\usepackage{CJK}
\usepackage{enumitem}
\usepackage{booktabs}
\usepackage{microtype}
\usepackage{arydshln}
\usepackage{amsfonts}
\usepackage{wasysym}
\usepackage{eufrak}
\usepackage{pifont}
\usepackage[normalem]{ulem}
\usepackage[T1]{fontenc}
\usepackage{tcolorbox}
\usepackage{algorithm,algorithmic}
\usepackage{xcolor,soul}
\usepackage{ellipsis}
\tcbuselibrary{breakable}

\definecolor{light_blue}{rgb}{0.80,0.85,1.0}
\definecolor{light_green}{rgb}{0.80,1.0,0.85}
\definecolor{light_red}{rgb}{1.0,0.85,0.80}
\definecolor{med_red}{rgb}{1.0,0.55,0.50}
\definecolor{light_orange}{rgb}{1.0,0.72,0.5}

\usepackage[utf8]{inputenc}

\begin{document}
\title{Resolving Knowledge Conflicts in Domain-specific Data Selection: A Case Study on Medical Instruction-tuning}
%
%
%

\author{
        Qihuang~Zhong,~\IEEEmembership{Member,~IEEE,}
        Liang~Ding,~\IEEEmembership{Member,~IEEE,}
        Fei Liao,
        Juhua~Liu,~\IEEEmembership{Member,~IEEE,}
        Bo~Du,~\IEEEmembership{Senior~Member,~IEEE,}
        and~Dacheng~Tao,~\IEEEmembership{Fellow,~IEEE}

\thanks{This work was supported in part by the National Key Research and Development Program of China under Grant 2023YFC2705700, in part by the National Natural Science Foundation of China under Grants 623B2076 and U23B2048, in part by the Science and Technology Major Project of Hubei Province under Grant 2024BAB046, in part by the Interdisciplinary Innovative Talents Foundation from Renmin Hospital of Wuhan University under Grant JCRCZN-2022-018, and in part by the Fundamental Research Funds for the Central Universities under Grand 2042024YXA002. The numerical calculations in this paper have been done on the supercomputing system in the Supercomputing Center of Wuhan University. \textit{Corresponding authors: Fei Liao, Juhua Liu.}}

\thanks{Qihuang Zhong, Fei Liao and Juhua Liu are with the Department of Gastroenterology, Renmin Hospital, Wuhan University, China. Qihuang Zhong, Juhua Liu and Bo Du are also with School of Computer Science, Wuhan University, China (e-mail: \{zhongqihuang, feiliao, liujuhua, dubo\}@whu.edu.cn).}

\thanks{Liang Ding is with the School of Computer Science, Faculty of Engineering, The University of Sydney, Australia (e-mail: liangding.liam@gmail.com).}

\thanks{Dacheng Tao is with the College of Computing \& Data Science at Nanyang Technological University, \#32 Block N4 \#02a-014, 50 Nanyang Avenue, Singapore 639798 (e-mail: dacheng.tao@ntu.edu.sg).}
}

\maketitle

\begin{abstract}
Domain-specific instruction-tuning has become the defacto standard for improving the performance of large language models (LLMs) in specialized applications, \textit{e.g.}, medical question answering. Since the instruction-tuning dataset might contain redundant or low-quality data, data selection (DS) is usually required to maximize the data efficiency. Despite the successes in the general domain, current DS methods often struggle to select the desired data for domain-specific instruction-tuning. One of the main reasons is that they neglect the impact of \textit{knowledge conflicts}, \textit{i.e.}, the discrepancy between LLMs' pretrained knowledge and context knowledge of instruction data, which could damage LLMs' prior abilities and lead to hallucination. To this end, we propose a simple-yet-effective Knowledge-aware Data Selection (namely \texttt{KDS}) framework to select the domain-specific instruction-tuning data that meets LLMs' actual needs. The core of \texttt{KDS} is to leverage two knowledge-aware metrics for quantitatively measuring knowledge conflicts from two aspects: context-memory knowledge alignment and intra-memory knowledge consistency. 
By filtering the data with large knowledge conflicts and sampling the high-quality and diverse data, \texttt{KDS} can effectively stimulate the LLMs' abilities and achieve better domain-specific performance.
Taking the medical domain as the testbed, we conduct extensive experiments and empirically prove that \texttt{KDS} surpasses the other baselines and brings significant and consistent performance gains among all LLMs. More encouragingly, \texttt{KDS} effectively improves the model generalization and alleviates the hallucination problem.
\end{abstract}

\section{Introduction}
\label{sec_intro}

\IEEEPARstart{W}{}hile large language models (LLMs)~\cite{openai2023gpt4,dubey2024llama,yang2024qwen2,liu2024deepseek,zhao2023survey} have showcased powerful capabilities in the general domain, they often struggle to handle the domain-specific tasks, \textit{e.g.}, medical question answering~\cite{labrak2024biomistral}. To enhance the performance of LLMs in these specialized applications, instruction-tuning~\cite{wei2021finetuned} on the specific domain is usually required. Different from traditional task-specific fine-tuning that relies on numerous training data, instruction-tuning only requires a relatively small dataset, as its goal is to align the LLMs' pretrained abilities in a desired direction~\cite{zhou2024lima}. Moreover, since the dataset might contain some undesired (\textit{e.g.}, low-quality and repetitive) samples, fine-tuning with the full dataset is usually sub-optimal. Hence, it is crucial to perform the domain-specific data selection (DS) for more effective instruction-tuning. 

\begin{figure}[t]
    \centering
    \includegraphics[width=0.48\textwidth]{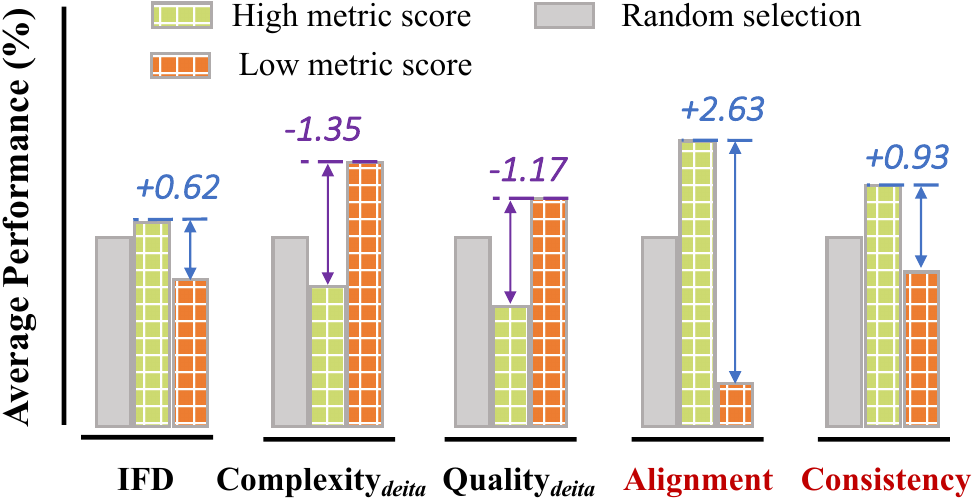}
    \caption{\textbf{Performance comparisons (\%) of different DS metrics}. Notably, ``IFD'' means the instruction-following difficulty~\cite{li2024quantity}, ``Complexity$_{deita}$'' and ``Quality$_{deita}$'' are from DEITA~\cite{liumakes}, and the metrics in red are ours. The y-axis denotes the average performance of tuned LLaMA models on several medical benchmarks, where the details are shown in Section~\ref{sec:experiments}.}
    \label{fig:metrics}
\end{figure}

Recently, in the general-domain instruction-tuning scenarios, some DS methods~\cite{chenalpagasus,li2024quantity,liumakes} have been proposed and achieved remarkable performance. Specifically, by using the heuristic automation (\textit{e.g.}, GPT-4 annotation) or manual selection, they can select high-quality and diverse data, which is beneficial to model training. However, in our preliminary experiments (as illustrated in Figure~\ref{fig:metrics}), we found that these methods might fail to select the desired data for domain-specific instruction-tuning. 
We conjecture that these DS methods are almost data-centric and overly focus on data quality and diversity, while neglecting whether the selected data meets LLMs' actual needs. This motivates us to explore a more effective model-adaptive DS method for domain-specific instruction-tuning scenarios.

Inspired by prior studies~\cite{manakul2023selfcheckgpt,xu2024knowledge,gekhman2024does,su2024conflictbank} related to LLM hallucination, we recognize that there is a critical issue in domain-specific instruction-tuning, \textit{i.e.}, \textbf{knowledge conflicts} between the LLMs' pretrained knowledge and the context knowledge of instruction-tuning training data. Since the world knowledge of LLMs is mainly learned during the pretraining stage and instruction-tuning fails to learn additional knowledge~\cite{ren2024learning}, enforcing the LLMs to align the contradictory domain knowledge through instruction-tuning would easily damage their prior abilities and lead to hallucination~\cite{gekhman2024does}. Thus, there raises a question: \textit{whether we can resolve the knowledge conflicts in domain-specific instruction-tuning and select the data desired by LLMs more effectively?}

To this end, we propose a knowledge-aware data selection framework (namely \texttt{KDS}) to tackle the knowledge conflicts and boost the LLMs' domain-specific performance. Specifically, \texttt{KDS} contains three key processes: \ding{182} \textit{multiple response generation}, \ding{183} \textit{knowledge-aware data scoring} and \ding{184} \textit{filtering and sampling}. First, in \ding{182}, we probe the LLMs' parametric pretrained knowledge by obtaining multiple candidate model responses for each question.
Then, in \ding{183}, to quantitatively evaluate the conflicts, we design two simple-yet-effective metrics: \textbf{knowledge alignment} and \textbf{knowledge consistency}. The former measures the fine-grained alignment between the LLM's responses and corresponding answers, while the latter focuses on the reference-free scenarios and uses the cluster-based semantic uncertainty to measure the consistency of LLM's multiple responses. Lastly, in \ding{184}, we further introduce two auxiliary strategies, \textit{i.e.}, quality filter and diversity filter, to ensure the quality and diversity of final selected data.

We take a representative domain-specific application (\textit{i.e.}, medical instruction-tuning) as the testbed, and evaluate the LLaMA3~\cite{dubey2024llama} and Qwen2.5~\cite{yang2024qwen2} models tuned with \texttt{KDS} on a variety of medical question-answering benchmarks. Extensive results show that our \texttt{KDS} not only surpasses the other DS methods by a clear margin, but also brings consistent and significant performance gains (up to \textbf{+2.56\%} average scores) across all LLMs. In-depth analyses prove that \texttt{KDS} can effectively improve data efficiency and multilingual generalization. More encouragingly, \texttt{KDS} alleviates the hallucination of tuned LLMs by bringing up to \textbf{+9.86\%} performance gains in the medical hallucination test.

To summarize, \textbf{our contributions} are as follows: 
\begin{itemize}
    \item We reveal that \textit{knowledge conflicts} are critical yet under-explored in domain-specific DS and propose a knowledge-aware DS (\texttt{KDS}) framework to resolve them.
    \item \texttt{KDS} design two simple-yet-effective metrics to quantitatively measure the knowledge conflicts from two aspects: context-memory knowledge alignment and intra-memory knowledge consistency.
    \item Extensive results on medical-domain benchmarks show that \texttt{KDS} outperforms the baselines by a clear margin and effectively improves the model generalization.
\end{itemize}

The rest of this paper is organized as follows. In Section~\ref{sec:related}, we briefly review the related works. In Section~\ref{sec:method}, we introduce our proposed framework in detail. Section~\ref{sec:experiments} reports and analyzes our experimental results, followed by the discussion in Section~\ref{sec:discussion}. Lastly, we conclude our study in Section~\ref{sec:conclusion}.

\section{Related Works}
\label{sec:related}
\subsection{Domain Adaptation of LLMs}
Recently, we have witnessed the great success of large language models (LLMs)~\cite{openai2023gpt4,liu2024deepseek,hu2023survey,jin2024large} in various of general-domain NLP tasks, such as machine translation~\cite{peng2023towards,zhang2023prompting},  mathematical reasoning~\cite{shao2024deepseekmath,zhong2024achieving}, sentiment analysis~\cite{zhang2022survey,zhong2023knowledge} and recommendation~\cite{fan2024recommender,zhang2025collm}. However, these LLMs still fall short in domain-specific applications, such as medical question-answering~\cite{labrak2024biomistral}. To improve the domain-specific performance of LLMs, there are two main ways: i) continual pretraining and ii) domain-specific instruction-tuning. Specifically, some previous works~\cite{chen2023meditron,labrak2024biomistral,yang2024zhongjing} attempt to continue pretraining the general-domain LLMs on the domain-specific corpus. Although this manner can effectively enrich the LLMs' domain knowledge and lead to better performance, it requires large-scale domain data and extensive training, which is costly and time-consuming, especially for some resource-constrained scenarios.

Hence, more researchers tend to use a straightforward and easy-to-implement approach, \textit{i.e.}, instruction-tuning~\cite{wei2021finetuned}. By fine-tuning LLMs with (relatively) small-scale instruction data, domain-specific instruction-tuning can unleash the power of LLMs in domain applications. For instance, in the medical domain, AlpaCare~\cite{zhang2023alpacare} creates a diverse medical instruction dataset by prompting GPT-4 and ChatGPT with a high-quality expert-curated seed set, and uses these data to fine-tune LLaMA models. ChatDoctor~\cite{li2023chatdoctor} collects a medical dataset of 100K patient-physician conversations from an online medical consultation website, and uses them to train a medical chat LLM. Despite its effectiveness, fine-tuning with a full instruction dataset is usually sub-optimal, as the dataset might contain some undesired (\textit{e.g.}, low-quality or repetitive) data. Training with these undesired data would easily lead to worse performance, or even hallucination problem~\cite{ji2023survey}. Hence, data selection (DS) for selecting the high-quality and desired subset appears to be crucial in domain-specific instruction-tuning. 


\begin{figure*}[ht]
    \centering
    \includegraphics[width=1\textwidth]{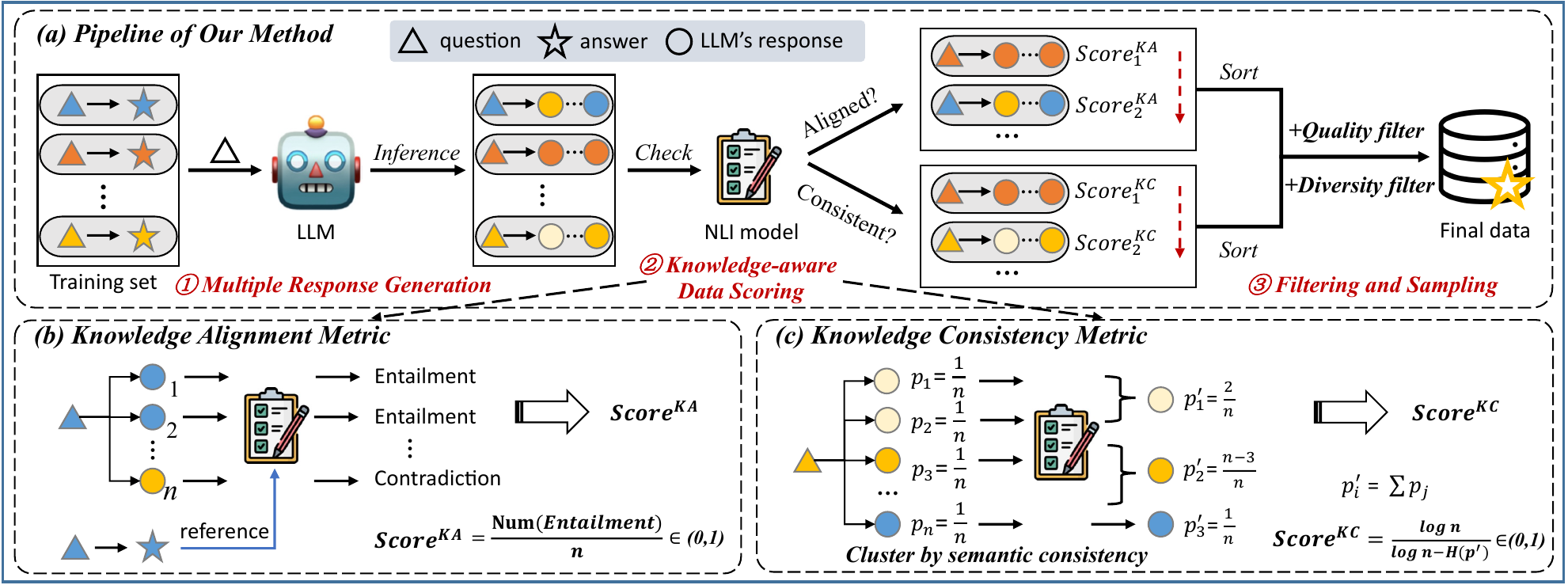}
    \caption{\textbf{Overview of our \texttt{KDS} framework}, which contains three processes: \ding{182} obtaining multiple responses of LLM for each question; \ding{183} scoring the data with the knowledge alignment and consistency metrics; \ding{184} filtering the low-quality and repetitive data, and sampling the final data. Notably, for ease of illustration, we only show a representative sample and simplified formulation in (b) and (c). $n$ denotes the number of responses for each question, $p_j=\frac{1}{n}$ is the assigned probability of $j$-th response and $p^{'}_{i}=\sum p_j$ is the sum of probabilities of $i$-th cluser.}
    \label{fig:method}
\end{figure*}

\subsection{Data Selection for Instruction-tuning}
In the general-domain instruction-tuning, many data-centric DS methods~\cite{chenalpagasus,liumakes,li2024quantity} have been proposed, which aim to select the high-quality and diverse data via heuristic methods (\textit{e.g.}, GPT-4 annotation) or manual selection. For instance, Chen~\textit{et~al.}~\cite{chen2023alpagasus} design a quality-oriented evaluation prompt to instruct the ChatGPT to assign the quality score for each instruction-response pair, and select the high-score data as the final training set. Li~\textit{et~al.}~\cite{li2024quantity} propose an instruction-following difficulty (IFD) metric to evaluate the difficulty of instruction data, where the high-score samples are considered as target data. Liu~\textit{et~al.}~\cite{liumakes} focus on the data quality and complexity, and train some LLM-based scorers to automatically evaluate the quality and complexity of instruction data. By training with the selected high-quality data, LLMs can avoid the negative effects of noisy data and achieve better performance.

Despite their remarkable performance in the general domain, these DS methods might fall short in selecting desired data for domain-specific instruction-tuning. Most of these methods are data-centric, while neglecting the actual needs of LLMs. Specifically, since the domain-specific instruction-tuning is more knowledge-intensive and contains rich professional knowledge that has not been learned during the LLMs' pretraining, the selected data could contain some knowledge that contradicts LLMs' internal knowledge, which is referred to \textit{knowledge conflict} problem. Enforcing LLMs to learn this conflict knowledge through instruction-tuning often leads to negative effects, \textit{e.g.}, catastrophic forgetting of prior knowledge, or hallucination problem~\cite{ren2024learning,gekhman2024does}.



More recently, some previous works~\cite{xu2024knowledge,wang2023resolving,manakul2023selfcheckgpt,zhao2024knowing} attempt to explore and resolve the knowledge conflict problem in LLM applications, such as RAG~\cite{jin2024tug} and factual reasoning~\cite{yu2023characterizing}. However, in the instruction-tuning field, there are only a few works~\cite{ren2024learning,ding20243ds,gekhman2024does} involving analyzing and resolving this problem, where two studies are most relevant to ours. Specifically, to detect the knowledge conflict in instruction data, Ren~\textit{et~al.}~\cite{ren2024learning} first employ in-context learning (ICL)~\cite{brown2020language} to probe LLMs' internal knowledge and determine whether it conflicts with the training data. For alleviating the negative effect of knowledge conflict, Ding~\textit{et~al.}~\cite{ding20243ds} design a simple prompt to instruct the LLMs to determine whether the context knowledge of instruction data is aligned with LLMs' internal knowledge, and propose to filter the unfamiliar data. Although achieving remarkable performance, these two methods still have some shortcomings. On the one hand, the proposed conflict detection methods are simply based on ICL or prompt engineering, which is sensitive to few-shot examples and might introduce bias into the results~\cite{min2022rethinking,ye2024empirical}. On the other hand, the method~\cite{ren2024learning} mainly focuses on the multiple-choice QA settings and might fall short in free-style generation tasks. 

To this end, we propose a knowledge-aware DS framework (\texttt{KDS}) to effectively resolve the knowledge conflicts in the domain-specific instruction-tuning scenarios. Specifically, instead of solely relying on the ICL or prompt engineering methods, \texttt{KDS} first obtains multiple candidate model responses, and then automatically evaluates the knowledge conflicts from two aspects: 1) context-memory knowledge alignment and 2) intra-memory knowledge consistency. The conflict detecting method of \texttt{KDS} is flexible and can be applied to various free-style generation scenarios. Thus, we believe that our work is valuable and could promote more related research.

\begin{table*}[htb]
    \centering
    \begin{tabular}{p{0.98\linewidth}}
        \toprule
\textbf{\textit{Low Knowledge Alignment}} \\ \hdashline 
\textbf{Question:} In which part of the nephron does approximately 20\% of potassium (K+) reabsorption occur? \\
\textbf{Answer:} Approximately 20\% of K+ reabsorption occurs in the {\sethlcolor{light_blue}\hl{thick ascending limb (loop of Henle)}}. \\ \vspace{-1mm}
\textbf{\textit{Multiple Responses:}} \\
\textbf{1.} According to physiological knowledge, approximately 20\% of potassium (K+) reabsorption occurs in the {\sethlcolor{light_red}\hl{proximal convoluted tubule (PCT)}} of the nephron. \\
\textbf{2.} According to physiological and anatomical knowledge, approximately 20\% of potassium (K+) reabsorption occurs in the {\sethlcolor{light_red}\hl{proximal convoluted tubule (PCT)}} of the nephron. \\
\textbf{3.} According to physiological knowledge, approximately 20\% of potassium (K+) reabsorption occurs in the {\sethlcolor{light_red}\hl{distal convoluted tubule (DCT)}} of the nephron.  \\
\textbf{4.} According to physiological and anatomical knowledge, approximately 20\% of potassium (K+) reabsorption occurs in the {\sethlcolor{light_red}\hl{distal convoluted tubule (DCT)}} of the nephron.  \\
\textbf{5.} A question about kidney physiology! According to various physiological and anatomical sources, approximately 20\% of potassium (K+) reabsorption occurs in the {\sethlcolor{light_red}\hl{proximal convoluted tubule (PCT)}} of the nephron.\\ \ldots  \\ \midrule
\textbf{\textit{High Knowledge Alignment}} \\ \hdashline 
\textbf{Question:} Which cranial nerve provides innervation to the levator palpebrae superioris muscle? \\
\textbf{Answer:} The levator palpebrae superioris muscle is innervated by the {\sethlcolor{light_blue}\hl{oculomotor nerve (cranial nerve III)}}. \\ \vspace{-1mm}
\textbf{\textit{Multiple Responses:}} \\
\textbf{1.} Truthfully... The cranial nerve that provides innervation to the levator palpebrae superioris muscle is the {\sethlcolor{light_green}\hl{III (Oculomotor) nerve}}! \\
\textbf{2.} The cranial nerve that provides innervation to the levator palpebrae superioris muscle is the **{\sethlcolor{light_green}\hl{oculomotor nerve (III)}}**. \\
\textbf{3.} Truthfully, the answer is the {\sethlcolor{light_green}\hl{oculomotor nerve (cranial nerve III)}}.  \\
\textbf{4.} The answer is: The cranial nerve that provides innervation to the levator palpebrae superioris muscle is the {\sethlcolor{light_green}\hl{Third Cranial Nerve (Oculomotor Nerve)}}, specifically its superior division.  \\
\textbf{5.} The cranial nerve that provides innervation to the levator palpebrae superioris muscle is the {\sethlcolor{light_green}\hl{oculomotor nerve (CN III)}}.\\ \ldots  \\
\bottomrule
    \end{tabular}
    \caption{
    \textbf{Examples of medical instruction data with low/high \textit{KA} scores}. For ease of illustration, we only present 5 of 10 model responses for each question. Notably, the key information is highlighted, where light blue denotes reference answers, light red denotes wrong responses and light green denotes right responses.
    }
    \label{tab:knowledge_alignment}
\end{table*}

\section{Method}
\label{sec:method}

\subsection{Task Formulation}
Given a base LLM $\mathcal{M}_{intial}$ 
that has been trained in the general-domain SFT corpus and has the basic instruction-following ability, the task of domain-specific DS aims to select an optimal training subset for maximizing the LLM's target domain performance, \textit{i.e.,} medical question-answering in our study. Let $\mathcal{D}$ denotes the full training set, containing $n$ queries $\mathcal{Q}=\{q_1,q_2,...,q_n\}$ and their corresponding answers $\mathcal{A}=\{a_1,a_2,...,a_n\}$, we employ the DS methods to select a desired training subset  $\mathcal{S} \subseteq \mathcal{D}$ of size $k$, where $k$ is the data budget. Lastly, we fine-tune the $\mathcal{M}_{intial}$ on $\mathcal{S}$ and obtain the final domain-specific LLM $\mathcal{M}_{final}$.

\subsection{Knowledge-aware Data Selection}


To tackle knowledge conflicts, the most important thing is to quantitatively measure them. According to Xu~\textit{et~al.}~\cite{xu2024knowledge}, there are two main types of conflicts: context-memory and intro-memory conflicts. The former refers to the discrepancy between pretrained knowledge of $\mathcal{M}_{intial}$ and context knowledge in $\mathcal{D}$, while the latter refers to the divergence of multiple responses of $\mathcal{M}_{intial}$ for the same question. Hence, in \texttt{KDS}, we first design two metrics to measure both types of conflicts, respectively. Then, considering the importance of data quality and diversity, we further introduce two auxiliary strategies to filter the low-quality and repetitive data. The overview of \texttt{KDS} is shown in Figure~\ref{fig:method}, containing three core processes:

\ding{182} \textbf{Multiple Response Generation}: 
To detect knowledge conflicts, we first need to explicitly probe the LLM's parametric pretrained knowledge. A straightforward way is to feed the question $q_i$ into $\mathcal{M}_{intial}$ and obtain its response, as the model's response to a question can usually reflect its internal knowledge. Furthermore, considering the instability of LLM's output, we are inspired by self-consistency~\cite{wangself}, and propose to sample a set of candidate responses $\{r^i_1, r^i_2,..., r^i_m\}$ from the LLM's decoder, where $m$ is the number of responses. By doing so, we can fully explore LLM's internal knowledge and explicitly obtain more comprehensive knowledge representation. In practice, for response generation, we set the temperature to 0.7 and sample $m=10$ responses for each question.

\ding{183} \textbf{Knowledge-aware Data Scoring}:
Given the reference answers and LLM's multiple responses, we design two metrics to measure the knowledge conflicts: \textit{\textbf{Knowledge Alignment}} (termed \textit{KA}) and \textit{\textbf{Knowledge Consistency}} (termed \textit{KC}). The primary intuition of \textit{KA} is that, for a question, if LLM's response contradicts the answer, the LLM does not know the knowledge for the question, \textit{i.e.}, there is a knowledge conflict. Specifically, 
inspired by prior studies related to textual entailment~\cite{farquhar2024detecting,kuhnsemantic}, 
we use an external NLI model to judge the relationships between LLM's responses and answers. The NLI model can determine whether the model response contradicts the reference answer according to the similarity of sentence representations. The calculation of \textit{KA} can be formulated as:
\begin{equation}
    Score^{KA}_i=\frac{\sum_{j=1}^{m}\mathbb{I} (\text{NLI}(r^i_j, a_i) = \text{entailment})}{m},
    \label{eq:ka}
\end{equation}
where $Score^{KA}_i$ is the \textit{KA} score for $i$-th data, $\text{NLI}(\cdot)$ is the inference results of NLI model, classified into either entailment/neutral/contradiction. For a better understanding of \textit{KA} metric, we show some comparative examples of high and low metric scores in Table~\ref{tab:knowledge_alignment}. As seen, our method can indeed distinguish the samples with high knowledge conflicts.

On the other hand, since the answers in $\mathcal{A}$ might be low-quality, misleading or even unavailable in some scenarios, the \textit{KA} would not work. Therefore, we further design a reference-free \textit{KC} metric, which focuses on the intra-memory conflict. Intuitively, if $\mathcal{M}_{intial}$ is not familiar with the knowledge of $q_i$, it shows a high uncertainty and may yield divergent responses. That is, higher uncertainty between multiple responses generally refers to larger knowledge conflicts. Here, to quantitatively evaluate the uncertainty, we 
propose a cluster-based knowledge consistency metric. Let $p_{i_j}=\frac{1}{m}$ be the uniform probability for $j$-th response of $\mathcal{M}_{intial}$, we cluster the responses with similar knowledge by using the NLI model for semantic matching. Specifically, if two responses are determined as ``entailment'', we treat them as the same cluster. Then, we calculate the entropy of clusters as the $Score^{KC}$, which is formulated as:
\begin{align}
    \nonumber &p^{'}_{i_t}=\sum_{j \in cluster} p_{i_j},\quad H(p^{'}_i) = -\sum_{t=1}^{c_i} p^{'}_{i_t} \log{p^{'}_{i_t}}, \\
    &Score^{KC}_i=\frac{\log{m}}{\log{n}-H(p^{'}_i)} \in (0, 1),
    \label{eq:kc}
\end{align}
where $p^{'}_{i_t}$ is the sum of probabilities of $t$-th cluster for $i$-th training data, $c_i \leq m$ is the number of clusters for $i$-th data, $H(p^{'}_i)$ is the entropy for $i$-th data and $\log{m}$ is the entropy upperbound. Lastly, we can sort the full $\mathcal{D}$ by using the individual $Score^{KA}$ and $Score^{KC}$, or the combination ``$Score^{KA}$+$Score^{KC}$'' as the metric.

\ding{184} \textbf{Filtering and Sampling}:
Based on the aforementioned scores, we can sort the full $\mathcal{D}$ and simply select the Top-$k$ samples as the final data. However, since the training set $\mathcal{D}$ might contain some low-quality and repetitive samples, which are harmful for instruction-tuning, we further introduce two auxiliary strategies, \textit{i.e.}, \textit{quality filtering} and \textit{diversity filtering}, to ensure the data quality and diversity. 

For the quality filtering, different from previous works~\cite{chenalpagasus,liumakes} that require additional LLM-based quality scorers, we propose to fully leverage the $\mathcal{M}_{intial}$'s powerful in-context learning and judgment capabilities~\cite{gu2024survey} for quality assessment. In practice, we design a quality-oriented prompt to instruct the $\mathcal{M}_{intial}$ itself to rate the data from 0 to 5, and filter the low-quality data with scores below the threshold $\tau$. The quality-oriented prompt is shown as follows:

\begin{tcolorbox}
[colback=lightgray!20,colframe=darkgray!80,title= Quality Scoring Prompt]
\label{tab:quality_prompt}
You are a fair and professional medical AI assistant. Your task is to rate according to the quality of the response to the instruction and the input. Each response receives a score on a scale of 0 to 5, where a higher score indicates a higher level of quality. Please directly output a single line containing the value indicating the scores. 
\newline
\newline
Instruction: \texttt{<instruct>}
\newline
Input: \texttt{<question>}
\newline
Response: \texttt{<answer>}

\end{tcolorbox}

\begin{figure}[h]
    \includegraphics[width=0.49\textwidth]{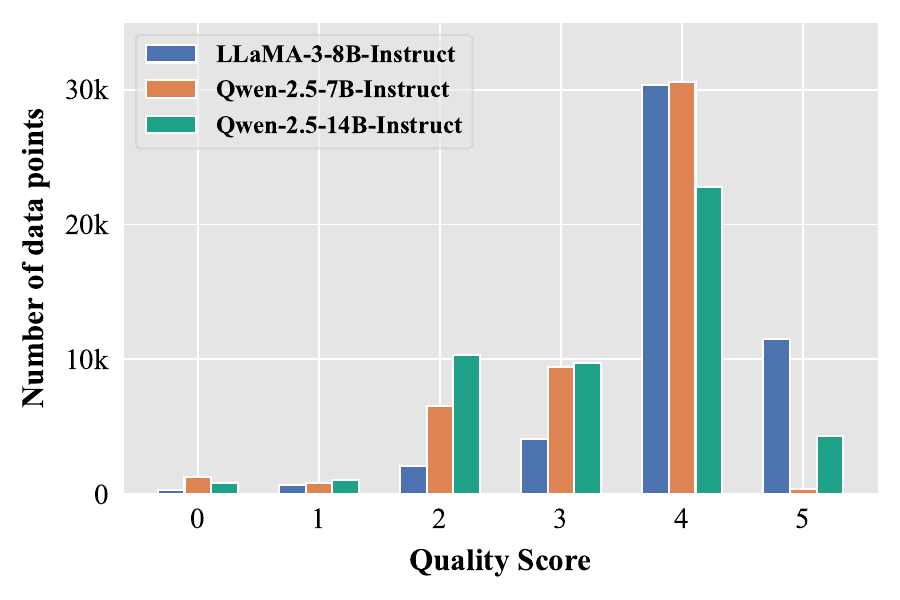}
    \caption{\textbf{Distributions of quality score} measured by different base LLMs. The x-axis denotes the measured quality score, and the y-axis denotes the number of data points.}
    \label{fig:quality_scores}
\end{figure}

To have a close look, in Figure~\ref{fig:quality_scores}, we illustrate the distributions of quality scores on the full training set measured by different base LLMs. As seen, the quality scores of different LLMs generally show a normal distribution, indicating that LLMs can measure the quality of medical data. 

On the other hand, towards the diversity filtering, inspired by prior work~\cite{liumakes}, we first convert all data into sentence embeddings using the BGE-m3\footnote{https://huggingface.co/BAAI/bge-m3} model~\cite{chen2024bge} and calculate the cosine distance between the data and its nearest neighbor in the current subset. The data with cosine distance below the threshold $\lambda$ will be filtered. This process is iterative and stops until the size of current subset exceeds the data budget $k$. Overall, the pipeline of \texttt{KDS} is shown in Algorithm~\ref{alg:kds}.

\begin{algorithm}[t]
  \footnotesize
	\renewcommand{\algorithmicrequire}{\textbf{Input:}}
	\renewcommand{\algorithmicensure}{\textbf{Output:}}
	\caption{Knowledge-aware Data Selection}
	\label{alg1}

	\begin{algorithmic}[1]
            \STATE \textbf{Input:} The full training dataset $\mathcal{D}=\{\mathcal{Q},\mathcal{A}\}$, base LLM $\mathcal{M}_{intial}$, data budget $k$, quality filter threshold $\tau$, diversity filter threshold $\lambda$
            \STATE \textbf{Output:} The selected subset $\mathcal{S}$
		\STATE Initialize Empty Dataset $\mathcal{S}$
		\FOR{Each sample $(q, a) \in \mathcal{D}$}
			\STATE Obtaining multiple responses of $\mathcal{M}_{intial}$ for $q$
			\STATE Calculating $Score^{KA}$ in Eq.~\ref{eq:ka} or $Score^{KC}$ in Eq.~\ref{eq:kc}
		\ENDFOR
		     \STATE Sorting $D$ with individual $Score^{KA}$ and $Score^{KC}$, or the combination ``$Score^{KA}$+$Score^{KC}$''       \STATE Getting the sorted Pool $D^{*}$

            \FOR{Each sample $(q, a) \in \mathcal{D}^{*}$}
			\STATE Obtaining quality score $s_q$ using the quality scoring prompt
			\STATE Obtaining the sentence embedding $emb(q,a)$ using the BGE-m3 model
            \STATE // $Cos(emb(q,a), \mathcal{S})$ denotes the cosine distance between $emb(q,a)$ and its nearest neighbor in $\mathcal{S}$ 
			\IF{$s_q \geq \tau$ and $Cos(emb(q,a), \mathcal{S}) < \lambda$} 
            \STATE $\mathcal{S} \leftarrow \mathcal{S} \cup \{(q,a)\} $ 
            
           \ELSE
            \STATE Continue
             \ENDIF
            \IF{$|\mathcal{S}|$ equals to $k$}
            \STATE Break
            \ENDIF
            \ENDFOR
	\end{algorithmic}
\label{alg:kds}
\end{algorithm}

\section{Experiments}
\label{sec:experiments}
\subsection{Experimental Setup}
\subsubsection{Tasks and Datasets}
In this study, we mainly evaluate our \texttt{KDS} in a representative domain-specific application, \textit{i.e.}, medical instruction-tuning. Since there is no standard medical instruction-tuning dataset, like the Alpaca~\cite{taori2023stanford} in the general domain, we construct a medical instruction-tuning dataset by collecting some existing tasks from the MedAlpaca~\cite{han2023medalpaca}. Notably, considering the inference budgets, we do not use the full MedAlpaca dataset (about 1.5 million data points) but select some sub-tasks of similar data size to Alpaca, containing \textit{Medical Flashcards}, \textit{Wikidoc}, and \textit{Wikidoc Patient Information}. These tasks contain rich professional and up-to-date medical knowledge, based on which we can better simulate the knowledge conflict problem in domain-specific instruction-tuning scenarios. After collecting the data, we randomly select 49,000 samples as the training dataset and use the other 495 samples as the held-out test (denoted as HoT).

\begin{table}[t]
\centering
\resizebox{0.49\textwidth}{!}{%
\begin{tabular}{lll}
\toprule
\textbf{Dataset} & \textbf{\#Type} & \textbf{\#Sample} \\
\midrule
MedMCQA~\cite{pal2022medmcqa} &multi-choice & 4,183 \\
MedQA-4options~\cite{jin2021disease} &multi-choice & 1,273 \\
PubmedQA~\cite{jin2019pubmedqa} &multi-choice & 500 \\
MMMLU-Medical~\cite{hendrycksmeasuring} & & \\
\quad - Anatomy (\textbf{Anatomy}) &multi-choice & 135 \\
\quad - Clinical-Knowledge (\textbf{Clinical}) &multi-choice & 265 \\
\quad - College-Biology (\textbf{Biology}) &multi-choice & 144 \\
\quad - College-Medicine (\textbf{Medicine}) &multi-choice & 173 \\
\quad - Medical-Genetics (\textbf{Genetics}) &multi-choice & 100 \\
\quad - Professional-Medicine (\textbf{Pro-Med}) &multi-choice & 272 \\ \hdashline
Long-form Medical QA~\cite{hosseini2024benchmark} &long-form QA & 400 \\
\bottomrule
\end{tabular}
}
\caption{\textbf{Tasks descriptions and statistic information} of all evaluation datasets in the main experiments.}
\label{tab:data}
\end{table}

To make a comprehensive evaluation, we further evaluate the tuned models on several out-of-domain (OOD) benchmarks. Specifically, four multiple-choice QA benchmarks (MedMCQA~\cite{pal2022medmcqa}, MedQA (4-option)~\cite{jin2021disease}, PubmedQA~\cite{jin2019pubmedqa} and MMLU-Medical~\cite{hendrycksmeasuring} and a long-form medical QA benchmark~\cite{hosseini2024benchmark} are used. For evaluation, we use the public \texttt{lm-evaluation-harness}\footnote{https://github.com/EleutherAI/lm-evaluation-harness} toolkit to measure the zero-shot accuracy of LLMs on multiple-choice QA benchmarks, and utilize the Rouge-L~\cite{lin2004rouge} as the metric for the HoT test set. For the long-form medical QA benchmark, we follow the original paper and employ the \textbf{LLM-as-a-Judge} as the metric. In practice, we use the GPT-4o-mini to judge from multiple aspects, covering \textit{Correctness}, \textit{Helpfulness}, \textit{Harmfulness}, \textit{Reasoning}, and \textit{Efficiency}. The statistics of all evaluation datasets in the main experiments are shown in Table~\ref{tab:data}, where each task is described as follows:
\begin{itemize}
    \item \textbf{MedMCQA}~\cite{pal2022medmcqa}: It consists of 4-option multiple-choice QA samples from the Indian medical entrance examinations (AIIMS/NEET). This dataset covers 2.4K healthcare topics and 21 medical subjects. We use the validation set with 4,183 questions for evaluation.
    \item \textbf{MedQA}~\cite{jin2021disease}: This dataset consists of questions and corresponding 4-option or 5-option answers in the style of the US Medical License Exam (USMLE). We follow prior work~\cite{chen2023meditron} and use the 4-option MedQA with 1,273 samples as the evaluation set.
    \item \textbf{PubmedQA}~\cite{jin2019pubmedqa}: It consists of 200K artificially created multiple-choice QA samples and 1K expert-labeled samples. Given a PubMed abstract as context and a question, LLM needs to predict a yes, no, or maybe answer. Similar to \cite{singhal2023large}, we use the 500 test samples for evaluation.
    \item \textbf{MMLU-Medical}~\cite{hendrycksmeasuring}: MMLU is a comprehensive benchmark, including exam questions from 57 subjects, where each subject contains 4-option multiple-choice QA samples. Similar to prior works~\cite{singhal2023towards}, we select 6 subjects that are most relevant to medical and clinical knowledge: Anatomy, Clinical-Knowledge, College-Biology, College-Medicine, Medical-Genetics, and Professional-Medicine.
    \item \textbf{Long-form Medical QA}~\cite{hosseini2024benchmark}: It is a new publicly available medical benchmark of real-world consumer medical questions with long-form answer evaluation, annotated by medical doctors. For the evaluation criteria, it instructs the LLMs to perform the pairwise comparisons using a fine-grained annotation scheme, covering \textit{Correctness}, \textit{Helpfulness}, \textit{Harmfulness}, \textit{Reasoning}, \textit{Efficiency}, and \textit{Bias}. In our experiments, we found that almost all models exhibit similar bais performance. Thus, we ignore the \textit{Bias} and use the other criteria for evaluation.
\end{itemize}

\begin{table*}[ht]
\setlength{\tabcolsep}{6pt}
\resizebox{\textwidth}{!}{%
\begin{tabular}{lccccccccccc}
\toprule
\multicolumn{1}{c}{\multirow{2}{*}{\textbf{Method}}} & \multirow{2}{*}{\textbf{HoT}} &\multirow{2}{*}{\textbf{MedMCQA}} & \multirow{2}{*}{\textbf{MedQA}} & \multirow{2}{*}{\textbf{PubmedQA}} & \multicolumn{6}{c}{\textbf{MMLU-Medical}} & \multirow{2}{*}{\textbf{Avg.}} \\ \cmidrule(lr){6-11}
\multicolumn{1}{c}{} & & & &  & \textbf{Anatomy} & \textbf{Clinical} & \textbf{Biology} & \textbf{Medicine} & \textbf{Genetics} & \textbf{Pro-Med} & \\ \midrule \midrule
\multicolumn{12}{l}{\textit{Compared Results upon \textbf{LLaMA-3-8B-Instruct}}} \\ \hdashline
Base & 20.87 & 57.06 & 60.17 & 74.80 & 63.70 & 71.70 & 75.00 & 63.01 & {81.00} & \underline{75.00} & 47.41 \\
Full-SFT & 29.36 & 54.72  & 59.86  & 68.20  & 62.96  & 72.83  & 75.00  & 63.01  & 78.00  & 68.01  & 47.02 \\
Random & 29.47 & 56.75  & 60.49  & 68.40  & \underline{68.89}  & 73.21  & \textbf{78.47}  & \underline{65.32}  & 80.00  & 73.16  & 48.05 \\
Alpagasus & 27.78 & 56.90  & \underline{60.33}  & 71.40  & 65.93  & 73.96  & 77.08  & \underline{65.32}  & {81.00}  & 71.69  & 48.15 \\
IFD & 26.89 & 55.92  & 59.23  & \underline{75.60}  & 66.67  & 73.96  & \underline{77.78}  & 61.85  & 79.00  & 70.59  & 48.21  \\
DEITA & 28.69 & 55.10  & 58.92  & 73.60  & \textbf{69.63}  & 74.47 & \textbf{78.47} & 60.69 & 78.00 & 72.06 & 48.01 \\
3DS & 27.86 & 55.32 & 59.15 & 72.80 & 67.41 & \underline{75.09} & \textbf{78.47} & 63.58 & 80.00 & 72.43 & 47.99  \\
\rowcolor{gray!20} \texttt{KDS}-\textit{KA} & \textbf{31.64} & \textbf{57.42} & 59.23 & \textbf{76.60} & 65.93 & \textbf{75.47} & \underline{77.78} & \textbf{65.90} & \textbf{84.00} & \textbf{75.37} & \underline{49.83}  \\
\rowcolor{gray!20} \texttt{KDS}-\textit{KC} & {30.25} & {57.18} & \textbf{60.57} & 73.60 & 67.41 & \textbf{75.47} & \textbf{78.47} & 64.74 & \textbf{84.00} & 73.16 & {49.25} \\ 
\rowcolor{gray!20} \texttt{KDS}-\textit{KA}+\textit{KC} & \underline{31.09} &\underline{57.30} &60.09 &\textbf{76.60} &\textbf{69.63} &74.72 &\textbf{78.47} &\underline{65.32} &\underline{82.00} &74.63 &\textbf{49.87} \\ \midrule
\multicolumn{12}{l}{\textit{Compared Results upon \textbf{Qwen-2.5-7B-Instruct}}} \\ \hdashline
Base & 24.78 & 56.20 & {62.14} & 73.00 & \underline{72.07} & 77.36 & 86.11 & 67.63 & 83.00 & 76.47 & 48.87 \\
Full-SFT & \textbf{35.55} & \textbf{57.11} & 60.57 & 73.20 & 71.11 & 75.85 & 84.72  & 68.21 & 83.00 & 76.10 & 50.49 \\
Random & 34.56 & 55.65 & 60.80 & 73.00 & 67.41 & 76.60 & 83.33  & \textbf{69.36} & 82.00 & 76.47 & 49.98 \\
Alpagasus & 34.39 & 55.82 & 62.06 & 74.00 & 70.37 & \underline{78.49} & 85.42  & \textbf{69.36} & 85.00 & 76.47 & 50.63 \\
IFD & 30.61 & 52.81 & 60.49 & {75.40} & 68.15 & 77.74 & 84.72  & 68.21 & 81.00 & 76.47 & 49.23 \\
DEITA & 29.42 & 55.73 & 62.06 & 74.80 & 71.11 & 77.36 & 84.03  & 65.90 & 82.00 & 74.63 & 49.64 \\
3DS & 28.88 & 55.83 & 61.43 & 74.00 & 71.11 & \textbf{78.87} & \underline{86.81} & \underline{68.79} & 84.00 & 76.84 & 49.65 \\
\rowcolor{gray!20} \texttt{KDS}-\textit{KA} & {35.45} & 55.82 & 61.51 & \underline{75.60} & 71.11 & 78.11 & \textbf{87.50} & 68.21 & \underline{86.00} & \underline{77.21} & {51.07} \\
\rowcolor{gray!20} \texttt{KDS}-\textit{KC}  & 35.17 & \underline{56.42} & \underline{62.53} & 75.00 & 71.11 & \underline{78.49} & 84.72  & 68.21 & \textbf{87.00} & \textbf{79.04} & \underline{51.20} \\ 
\rowcolor{gray!20} \texttt{KDS}-\textit{KA}+\textit{KC} & \underline{35.30} & 56.04 & \bf 62.84 & \bf 76.20 & \bf 74.07 & 78.11 & {85.42} & 68.21 & \underline{86.00} & 76.47 & \bf 51.40 \\
\midrule
\multicolumn{12}{l}{\textit{Compared Results upon \textbf{Qwen-2.5-14B-Instruct}}} \\ \hdashline
Base & 24.03 & 63.61 & 69.84 & {78.20} & \underline{75.30} & \underline{83.77} & 89.58 & 75.72 & 88.00 & 83.46 & 53.05 \\
Full-SFT & \underline{36.63} & 62.90 & 69.05 & 76.60 & 72.59 & 83.02 & \textbf{90.97} & \textbf{78.03} & \textbf{91.00} & 83.82 & 54.74 \\
Random & 35.59 & 62.90 & 69.31 & 77.60 & \textbf{75.56} & \textbf{84.15} & 90.28 & 75.14 & 88.00 & 84.93 & 54.74 \\
Alpagasus & 35.86 & 63.50 & 69.78 & 77.80 & \textbf{75.56} & 82.64 & 89.58 & 75.72 & \underline{89.00}  & 85.03 & 54.98 \\
IFD & 35.07 & 63.33 & {69.99} & 77.80 & 73.33 & 82.26 & 89.58 & 75.14 & 88.00 & {85.66} & 54.75 \\
DEITA & 30.50 & 63.33 & 69.21 & \textbf{78.60} & 72.59 & 83.02 & 88.89 & 76.30 & \underline{89.00}  & 83.82 & 53.99 \\
3DS & 32.30 & 63.11 & 69.36 & 78.00 & \textbf{75.56} & 83.02 & 88.19 & 73.99 & \textbf{91.00} & 82.72 & 54.20 \\
\rowcolor{gray!20} \texttt{KDS}-\textit{KA} & 36.53 & {63.71} & \underline{70.46} & \textbf{78.60} & 74.07 & \underline{83.77} & \underline{90.28} & {77.46} & \underline{89.00}  & \textbf{86.76} & \underline{55.48} \\
\rowcolor{gray!20} \texttt{KDS}-\textit{KC} & \underline{36.74} & \textbf{63.88} & 69.76 & 77.80 & \textbf{75.56} & \underline{83.77} & 89.85 & 76.30 & \underline{89.00} & 85.29 & {55.25} \\
\rowcolor{gray!20} \texttt{KDS}-\textit{KA}+\textit{KC} & \bf 36.81 & \underline{63.78} & \bf 70.86 & \underline{78.40} & \bf 75.56 & 83.40 & 88.89 & \underline{77.49} & \bf 91.00 & \underline{86.40} & \bf 55.61 \\
\bottomrule

\end{tabular}
}
\caption{\textbf{Performance comparison (\%) on the held-out test (HoT) and multiple-choice medical QA benchmarks.} ``Avg.'' denotes the macro-average performance. Best results are in \textbf{bold}, and second-best results are \underline{underlined}.}
\label{tab:main}
\end{table*}

\subsubsection{Models}
We conduct extensive experiments on three widely-used LLMs across different model architectures and sizes, \textit{i.e.}, LLaMA-3-8B-Instruct~\cite{dubey2024llama}, Qwen-2.5-7B/14B-Instruct~\cite{yang2024qwen2}. Within our framework, we use the powerful \texttt{DeBERTa-v3-large-mnli}\footnote{https://huggingface.co/MoritzLaurer/DeBERTa-v3-large-mnli-fever-anli-ling-wanli} as the NLI model, and set the quality threshold $\tau$ to 3 and diversity threshold $\lambda$ to 0.9. We fine-tune the LLMs using the instruction data selected by different methods. The default data budget $k$ is set as 5,000. For model training, we fine-tune all LLMs with a batch size of 32 and a peak learning rate of 1e-4. The warm-up ratio is 0.1, and the maximum tokenizer length is 2,048. All models are trained with LoRA~\cite{hulora} for 3 epochs. We conduct all experiments on 8 NVIDIA A100 (40GB) GPUs. During inference, we set the temperature to 0 for reproducibility, and set the maximum output length to 256 tokens. 

\begin{figure*}[t]
    \centering
    \includegraphics[width=\textwidth]{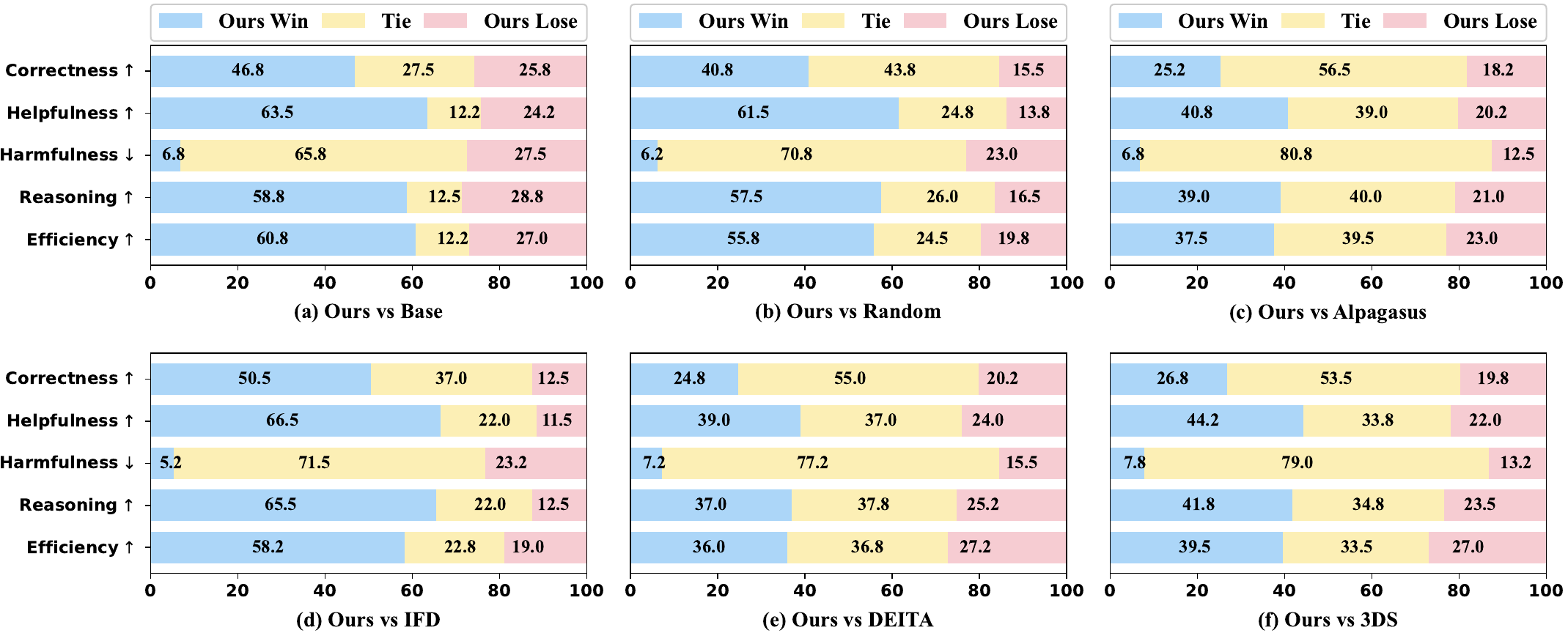}
    \caption{\textbf{Comparative winning rates (\%) of \texttt{KDS}-\textit{KA}+\textit{KC} v.s. other baselines on the long-form medical QA benchmark}~\cite{hosseini2024benchmark}. LLaMA-3-8B-Instruct is used as the base model, and GPT-4o-mini is used as the automated evaluator.}
    \label{fig:elo_all}
\end{figure*}

\subsubsection{Baselines}
We compare our proposed \texttt{KDS} method with a series of counterparts: \textit{Random}, \textit{Alpagasus}~\cite{chenalpagasus}, \textit{IFD}~\cite{li2024quantity}, \textit{DEITA}~\cite{liumakes} and \textit{3DS}~\cite{ding20243ds}. For reference, we also report the results of base models (\textit{Base}) and the models fine-tuned with the full training dataset (\textit{Full-SFT}). We re-implement the compared baselines following the original papers. The implementation of baselines is introduced as follows:
\begin{itemize}
    \item \textbf{Full-SFT}: We fine-tune the LLMs with the full instruction-tuning training dataset without using DS methods. This baseline is used to demonstrate the necessity of DS for domain-specific instruction-tuning.
    \item \textbf{Random}: We randomly sample 5K data from the instruction-tuning training dataset and fine-tune the LLMs with this data. This baseline is used as the vanilla DS.
    \item \textbf{IFD}~\cite{li2024quantity}: Following the original paper~\cite{li2024quantity}, we first calculate the Instruction Following Difficulty (IFD) scores for each data point of the instruction-tuning training dataset, and filter the data with IFD scores exceeding 1. Lastly, we sort the dataset based on IFD scores and select the top 5K data as the training subset.
    \item \textbf{Alpagasus}~\cite{chenalpagasus}: Chen~\textit{et~al.}~\cite{chenalpagasus} design a prompt to instruct the ChatGPT to score the data and select the high-score subset. In our implementation, we employ the same prompt and use the GPT-4o-mini as the automatic evaluator to score the data. After sorting the data based on the score, we select the top 5K data for training.
    \item \textbf{DEITA}~\cite{liumakes}: It aims to select the data via a quality scorer and a complexity scorer. In practice, we first score and sort the data by using the open-source LLaMA-based quality\footnote{https://huggingface.co/hkust-nlp/deita-quality-scorer} and complexity scorers\footnote{https://huggingface.co/hkust-nlp/deita-complexity-scorer}. Then, we use the recommended diversity-oriented method in~\cite{liumakes} to select the top 5K diverse data as the training corpus.
    \item \textbf{3DS}~\cite{ding20243ds}: It is the most relevant method to us, which also attempts to select the data that meets the LLMs' actual needs in the medical IT field. However, different from ours, it first filters irrelevant or redundant data via a prompt and uses three metrics (\textit{i.e.}, Instruction Understanding, Response Confidence, and Response Correctness) to select the appropriately challenging data. We use the same prompt and follow the recommendations in the original paper to select the final 5K samples.
\end{itemize}

\subsection{Compared Results}
The main results on HoT and multiple-choice medical QA are reported in Table~\ref{tab:main}, and the results of LLaMA models on long-form medical QA are illustrated in Figure~\ref{fig:elo_all}. From these results, we can find that:

\begin{figure*}[th]
    \centering
    \includegraphics[width=\textwidth]{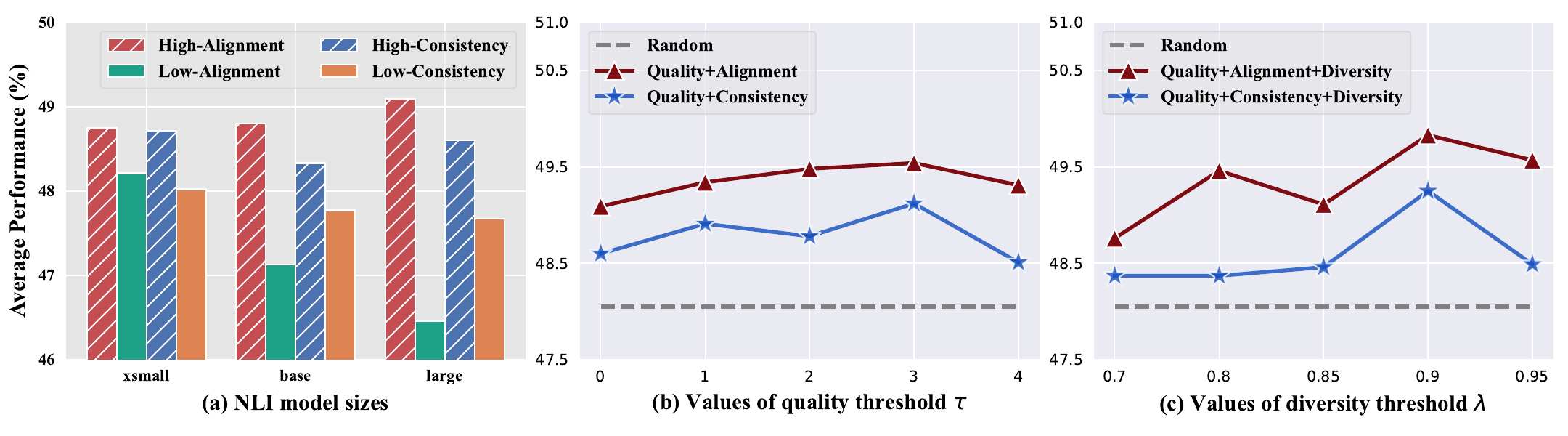}
    \caption{\textbf{(a) Effect of NLI models} with different model sizes, \textbf{(b) Parameter analysis of quality threshold $\tau$} and \textbf{(c) Parameter analysis of diversity threshold $\lambda$}. Notably, we use the LLaMA-3-8B-Instruct as the base model and report the average performance of HoT and multiple-choice QA benchmarks. }
    \label{fig:all_analysis}
\end{figure*}

\subsubsection{\texttt{KDS} surpasses the previous DS strategies by a clear margin} As seen, ``Full-SFT'' and ``Random'' usually perform poorly and even worse than the original base model. For instance, in the LLaMA models of Table~\ref{tab:main}, the average performance of base model is 47.41, while the ``Full-SFT'' is 47.02. These results prove the necessity of carefully-designed DS during the domain-specific adaptation of LLMs. Moreover, by comparing the performance of different DS methods, it can be found that the previous DS methods often struggle to improve the performance for medical question-answering, where the best-performing IFD method can only bring an average performance improvement of 0.80 in the LLaMA models. This indicates that overly pursuing data quality and diversity, while ignoring the knowledge conflicts problem, is sub-optimal for domain-specific instruction-tuning.

In contrast, by addressing the knowledge conflict problem, our \texttt{KDS} can achieve significantly better performance than other counterparts. Specifically, in the LLaMA models, \texttt{KDS} using \textit{KA} and \textit{KC} metrics can bring an average performance gain of 2.42 and 1.84, respectively. Moreover, by combining both metrics, \texttt{KDS} can bring further performance improvement. These results confirm our motivation in Section~\ref{sec_intro}.

\subsubsection{\texttt{KDS} brings consistent and significant performance gains among all model sizes and types} By comparing the performance of different base models in Table~\ref{tab:main}, we see that our \texttt{KDS} not only achieves remarkable performance on the LLaMA models, but also brings significant performance gains on the Qwen models. Specifically, compared to the base models, \texttt{KDS} achieves up to \textbf{+2.46\%}, \textbf{+2.53\%} and \textbf{+2.56\%} average gains for the LLaMA-3-8B-Instruction, Qwen-2.5-7B/14B-Instruct models, respectively. These results show that \texttt{KDS} method can be applied to various LLMs with different model scales and types, confirming its effectiveness and universality.

\subsubsection{\texttt{KDS} effectively improves the long-form QA performance}
In addition to above multiple-choice QA results, we also evaluate the tuned models on the long-form medical QA benchmark~\cite{hosseini2024benchmark} and illustrate the results in Figure~\ref{fig:elo_all}. Specifically, we show the winning rates of our method (\texttt{KDS}-\textit{KA}+\textit{KC}) against other baseline DS methods in the figure. The LLaMA3-8B-Instruct is used as the base model in this experiment.
As seen, compared to other baseline methods, the model trained with our \texttt{KDS} method can achieve better performance in terms of correctness, helpfulness, reasoning and efficiency, and reduce the harmfulness effectively. That is, \texttt{KDS} can not only bring better multiple-choice medical QA performance, but is also beneficial to long-form medical QA tasks.

\begin{table*}[t]
\centering
\resizebox{\textwidth}{!}{%
\begin{tabular}{lccccccccccc}
\toprule
\multicolumn{1}{c}{\multirow{2}{*}{\textbf{Method}}} & \multirow{2}{*}{\textbf{HoT}} &\multirow{2}{*}{\textbf{MedMCQA}} & \multirow{2}{*}{\textbf{MedQA}} & \multirow{2}{*}{\textbf{PubmedQA}} & \multicolumn{6}{c}{\textbf{MMLU-Medical}} & \multirow{2}{*}{\underline{\textbf{Avg.}}} \\ \cmidrule(lr){6-11}
\multicolumn{1}{c}{} & & & &  & \textbf{Anatomy} & \textbf{Clinical} & \textbf{Biology} & \textbf{Medicine} & \textbf{Genetics} & \textbf{Pro-Med} & \\ \midrule \midrule
Base & 20.87 & 57.06  & 60.17  & 74.80  & 63.70  & 71.70 & 75.00  & 63.01  & {81.00}  & {75.00}  & \underline{47.41} \\
Random & 29.47 & 56.75   & 60.49   & 68.40   & {68.89}   & 73.21   & {78.47}   & 65.32   & 80.00   & 73.16   & \underline{48.05} \\
\texttt{KDS} \textit{-w/o KA+KC} & 30.24	&56.9	&60.02	&70.60	&68.15	&73.58	&79.17	&65.90	&83.00	&70.22	&\underline{48.52} \\
\midrule
 \texttt{KDS}- \textit{KA} (Ours) &{31.64} & {57.42}  & 59.23  & {76.60}  & 65.93  & {75.47}  & {77.78}  & {65.90}  & {84.00}  & {75.37}  & \bf \underline{49.83} \\ \hdashline
\quad \textit{-w/o quality} & 32.17 & 56.87  & 59.78  & 75.20  & 65.93  & 76.23  & 77.08  & 65.90  & 83.00  & 73.16  & \underline{49.60} \\
\quad \textit{-w/o diversity} & 31.02 & 57.02  & 60.09  & {76.40}  & 66.67  & 72.83  & 77.78  & 63.58  & 81.00  & 74.26  & \underline{49.54} \\
\quad \textit{-w/o quality+diversity} & 31.51 & 56.39  & 60.41  & 72.80  & 65.93  & 74.72  & {79.86}  & 63.58  & {82.00}  & 74.63  & \underline{49.09} \\
\midrule
 \texttt{KDS}- \textit{KC} (Ours) & {30.25} & {57.18}  & {60.57}  & 73.60  & {67.41}  & {75.47}  & {78.47}  & 64.74  & {84.00}  & 73.16  & \bf \underline{49.25} \\ \hdashline
\quad \textit{-w/o quality}& 31.57 & 56.83  & 60.25  & 74.20  & 65.93  & 74.34  & 79.17  & 65.90  & 80.00  & 74.26  & \underline{49.35} \\
\quad \textit{-w/o diversity}& 30.13 & 57.04  & 60.02  & 75.60  & {67.41}  & 73.58  & 72.22  & 65.32  & 80.00  & 73.16  & \underline{49.12} \\
\quad \textit{-w/o quality+diversity} & 28.80 & 56.85  & {60.96}  & 71.80  & {67.41}  & 75.09  & 74.31  & {67.63}  & {82.00}  & 72.79  & \underline{48.60} \\
\bottomrule
\end{tabular}
}
\caption{\textbf{Ablation study on the different strategies}. Notably, LLaMA-3-8B-Instruct is used as the base model in this study.}
\label{tab:ablation}
\end{table*}

\subsection{Ablation Study}
\label{sec:ablation}
Here, we gradually investigate the effect of each important component of our \texttt{KDS}. Notably, we mainly use the LLaMA-3-8B-Instruct as the base model and report the average performance of HoT and multiple-choice QA benchmarks in this part. To better investigate the effect of \textit{KA}/\textit{KC}, we use the individual metric in our \texttt{KDS}.

\subsubsection{Effect of data filter strategies} As mentioned in Section~\ref{sec:method}, to ensure the data quality and diversity, we additionally introduce quality-oriented and diversity-oriented data filtering strategies upon the \textit{KA} and \textit{KC} metrics. Here, to verify the effect of these strategies, we compare our full \texttt{KDS} with the following variants: a) ``\textit{-w/o KA+KC}'' removes the \textit{KA} and \textit{KC} metrics, and only uses the quality and diversity filtering strategies; b) ``\textit{-w/o quality}'' only removes the quality filtering strategy; c) ``\textit{-w/o diversity}'' only removes the diversity filtering strategy; d) ``\textit{-w/o quality\&diversity}'' removes both the quality and diversity filtering strategies. The contrastive results of LLaMA models are shown in Table~\ref{tab:ablation}, from which we find that: 1) compared to the full \texttt{KDS} method, ``\textit{-w/o KA+KC}'' leads to greatest performance degradation, indicating that our proposed knowledge conflict detection metrics play a core role in the \texttt{KDS} method; 2) removing each data filtering strategy will lead to performance degradation and the full \texttt{KDS} performs best. Specifically, when using \textit{KA} metric in \texttt{KDS}, ``\textit{-w/o quality}'' and ``\textit{-w/o diversity}'' lead to 0.23 and 0.29 average performance drops, respectively. These results prove the effectiveness of these data filtering strategies.

\subsubsection{Influence of NLI model sizes}
In \texttt{KDS}, we use an extra NLI model to determine whether LLMs' outputs are aligned with the references. Intuitively, a larger NLI model can achieve more accurate judgments and lead to better performance. To verify it, we conduct experiments by utilizing three different sizes of DeBERTa-based NLI models, \textit{i.e.}, \texttt{xsmall}, \texttt{base} and \texttt{large}. To better showcase its effect, we directly compare the performance between models trained with the high \textit{KA/KC} samples and those with low \textit{KA/KC} samples, where the larger performance gap indicates more accurate judgments. Figure~\ref{fig:all_analysis} (\textbf{a}) shows the contrastive results. As seen, larger NLI models indeed perform better in distinguishing the \textit{KA/KC} of samples, confirming our conjecture. For example, when using \texttt{xsmall} NLI model, the performance gap between high \textit{KA} and low \textit{KA} settings is 0.54. In contrast, the performance gap of \texttt{large} NLI model is 2.63. Thus, we choose to use the \texttt{DeBERTa-v3-large-mnli} as the NLI model.

\subsubsection{Impact of quality threshold $\tau$}
The threshold $\tau$, used to filter the low-quality data, is an important hyperparameter in \texttt{KDS}. In this study, we analyze its influence by evaluating the performance with different $\tau$ values, spanning from 0 to 4. The highest quality score is 5, and the corpus with a quality score of 5 might be less than 5K samples (as shown in Figure~\ref{fig:quality_scores}), thus we do not conduct experiments with $\tau=5$. Notably, since we are performing individual analyses of quality threshold, we do not use the diversity strategy here.
 Figure~\ref{fig:all_analysis} (\textbf{b}) illustrates the average results, in which we can find that: 1) increasing the $\tau$ from 0 to 3 brings consistent performance gains, indicating that filtering the low-quality data is necessary; 2) too large $\tau$ (\textit{i.e.}, 4) would lead to performance degradation, as many helpful samples might be ignored. \texttt{KDS} performs best with $\tau=3$, thus leaving as our default experimental settings.

\subsubsection{Impact of diversity threshold $\lambda$}
The factor $\lambda$, which is used to control the data diversity, also needs to be investigated. Here, we conduct contrastive experiments to analyze the effect of $\lambda$. Figure~\ref{fig:all_analysis} (\textbf{c}) illustrates the results of varied $\lambda$ ranging from 0.7 to 0.95. From the results, we observe that overemphasizing diversity (\textit{i.e.}, too large $\lambda$ values) may cause too many samples with high \textit{KA/KC} scores to be filtered, thus leading to significant performance drops. In contrast, appropriately reducing the $\lambda$ can achieve a better trade-off between model performance and data diversity. More specifically, the case of $\lambda=0.9$ performs best, and we thereby use this setting in our experiments.

\begin{figure}[t]
    \centering
    \includegraphics[width=0.48\textwidth]{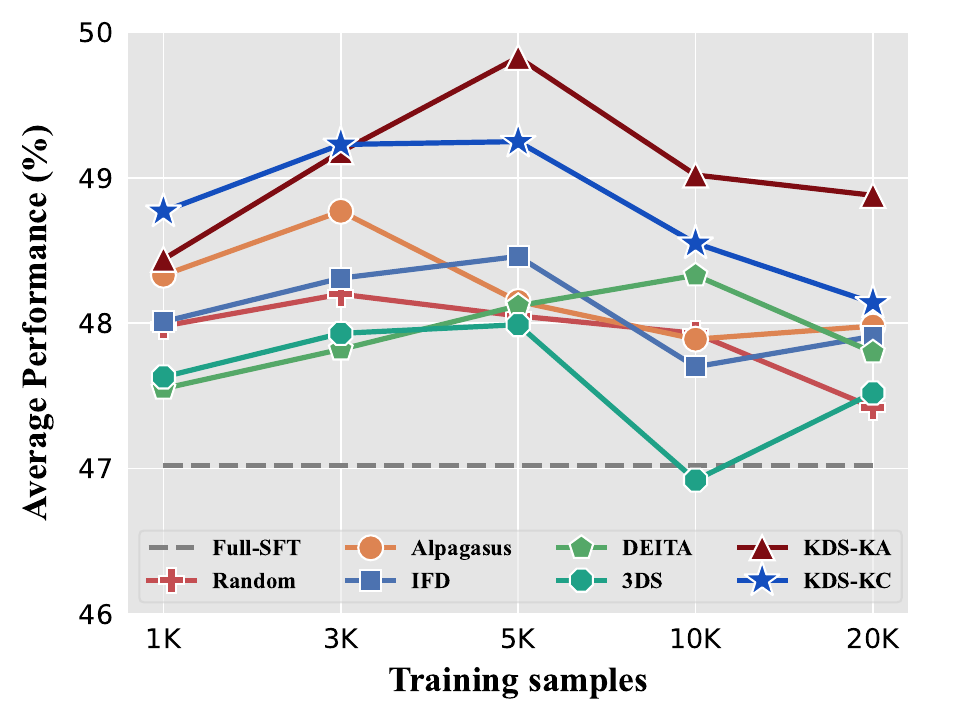}
    \caption{\textbf{Results at various training data scales}. We use the LLaMA-3-8B-Instruct as the base model.}
    \label{fig:data_scaling}
\end{figure}

\begin{figure*}[ht]
    \centering
    \includegraphics[width=\textwidth]{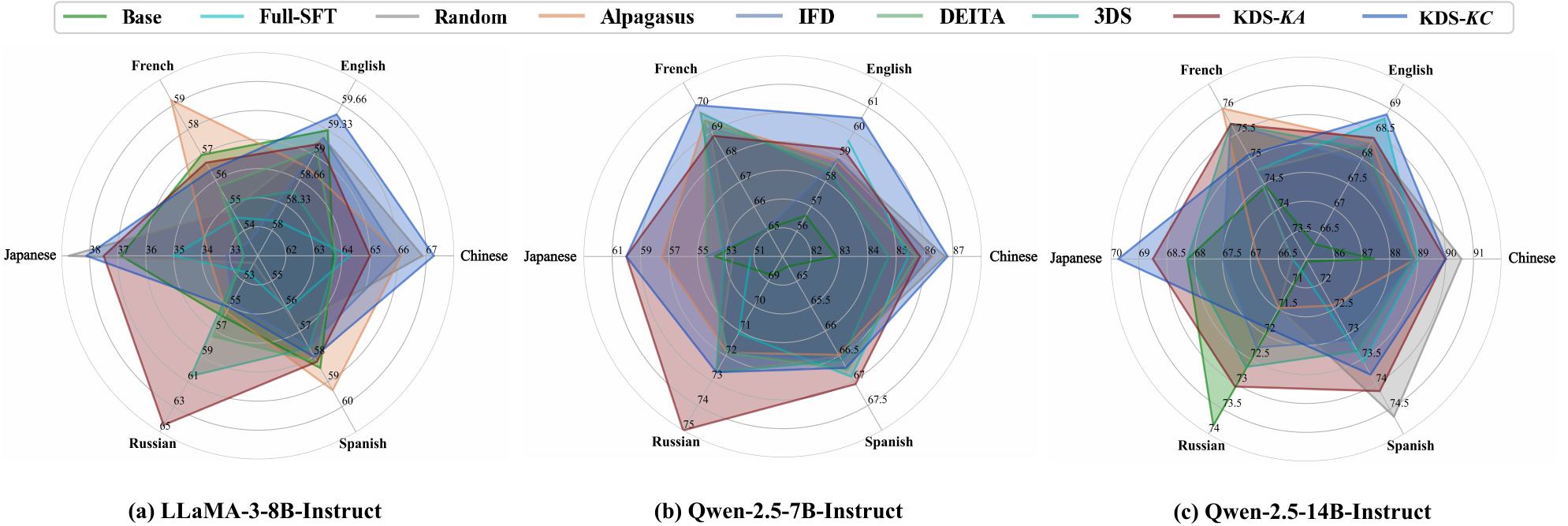}
    \caption{\textbf{Comparative results of LLMs tuned with different methods on the MMedBench}. MMedBench~\cite{qiu2024towards} is a multilingual medical multiple-choice QA benchmark across six primary languages: English, Chinese, Japanese, French, Russian, and Spanish.}
    \label{fig:mmedbench_all}
\end{figure*}

\section{Discussion}
\label{sec:discussion}
Here, we conduct further analyses to discuss: 1) whether \texttt{KDS} still works at other data scales, 2) whether \texttt{KDS} gains better model generalization, 3) whether \texttt{KDS} can be applied to more domain-specific scenarios, 4) the reliability of NLI model and 5) the efficiency of \texttt{KDS} pipeline. Lastly, we provide some case studies to qualitatively analyze the effectiveness of \texttt{KDS}.

\subsection{Does \texttt{KDS} still Work at other Data Scales?} 
In the above experiments, we mainly evaluate our \texttt{KDS} under the training data budget of 5K samples. Some readers may wonder whether \texttt{KDS} works in the other training settings. To verify it, we select varied numbers of samples using different data selection methods and use them to train the LLaMA3-8B model, respectively. The performance comparisons of different data selection methods are illustrated in Figure~\ref{fig:data_scaling}.
As seen, when the data scale is less than 5K, the model's performance stably improves as the number of samples increases. However, training with too many samples (\textit{e.g.}, 20K) will cause performance degradation, as the selected subset could contain a large amount of low-quality data and damage the training of models. It should be emphasized, among all data scales, our \texttt{KDS} can consistently outperform the other counterparts. More encouragingly, using only 1K training samples, our method can outperform other methods that use 5K samples. Specifically, when data budget $k$=1K, our \texttt{KDS}-\textit{KC} achieves an average performance of 48.77, which is higher than the best average performance (\textit{i.e.}, 48.46) of other DS methods with data budget $k$=5K. Takeaway: \textbf{\textit{our \texttt{KDS} method can effectively improve the data efficiency and work well at varied data scales.}}

\begin{table*}[ht]
\centering
\setlength{\tabcolsep}{13pt}
\resizebox{\textwidth}{!}{%
\begin{tabular}{lcccccccc}
\toprule
\multirow{2}{*}{\textbf{Method}} & \multicolumn{2}{c}{\textbf{Reasoning FCT}} & \multicolumn{2}{c}{\textbf{Reasoning Fake}} & \multicolumn{2}{c}{\textbf{Reasoning Nota}} & \multicolumn{2}{c}{\textbf{Average}} \\ \cmidrule(lr){2-3} \cmidrule(lr){4-5} \cmidrule(lr){6-7} \cmidrule(lr){8-9}
 & Acc & Score & Acc & Score & Acc & Score & Acc ($\Delta\uparrow$) & Score ($\Delta\uparrow$) \\ \midrule
Base & 43.28 & 54.86 & 74.76 & 12.72 & 35.18 & 35.80 & 51.07 & 34.46 \\ \hdashline
Random & 46.17 & 58.56 & 53.18 & 7.71 & 15.11 & -11.53 & 38.15 & 18.25 \\
Alpagasus & 45.74 & 60.69 & \underline{60.98} & \underline{9.52} & 18.37 & -3.84 & 41.70{$_{\uparrow3.55}$} & 22.12{$_{\uparrow3.85}$} \\
IFD & 47.45 & 64.73 & 51.24 & 7.26 & 17.81 & -5.18 & 38.83{$_{\uparrow0.68}$} & 22.27{$_{\uparrow4.02}$} \\
DEITA & 46.95 & 63.50 & 55.38 & 8.22 & 16.73 & -7.70 & 39.69{$_{\uparrow1.54}$} & 21.34{$_{\uparrow3.09}$} \\
3DS & 40.78 & 48.96 & \textbf{61.52} & \textbf{9.64} & 15.46 & -10.72 & 39.25{$_{\uparrow1.10}$} & 15.96{$_{\downarrow2.29}$} \\
\rowcolor{gray!20} \texttt{KDS}-\textit{KA} & \underline{48.17} & \underline{66.41} & 56.35 & 8.44 & \underline{21.98} & \underline{4.66} & \textbf{42.17}{$_{\uparrow\textbf{4.02}}$} & 26.50{$_{\uparrow8.25}$} \\
\rowcolor{gray!20} \texttt{KDS}-\textit{KC} & \textbf{48.90} & \textbf{68.14} & 51.24 & 7.26 & \textbf{23.79} & \textbf{8.94} & 41.31{$_{\uparrow3.16}$} & \textbf{28.11}{$_{\uparrow\textbf{9.86}}$} \\
\bottomrule
\end{tabular}
}
\caption{\textbf{Results of different tuned Qwen2.5-7B models on the medical hallucination evaluation benchmark, \textit{i.e.}, MedHalt}~\cite{pal2023med}. Notably, the subscript results refer to the performance differences against the ``Random'' baseline. The best results among all DS methods are in \textbf{bold}, where the second ones are \underline{underlined}.}
\label{tab:hallucination}
\end{table*}

\subsection{Does \texttt{KDS} Improve the Generalization?}
The instruction-tuning is known to improve the model generalization of LLMs~\cite{wei2021finetuned}. Intuitively, by selecting the high-quality data aligned with LLMs' prior knowledge, \texttt{KDS} can achieve smoother and more effective domain adaptation, thus resulting in better generalization. To verify it, we further analyze the effect of \texttt{KDS} from the following aspects:

\subsubsection{Multilingual Generalization} We evaluate the medical LLMs trained with different methods on the popular multilingual medical QA benchmarks, \textit{i.e.}, MMedBench~\cite{qiu2024towards}, and illustrate the comparative results in Figure~\ref{fig:mmedbench_all}. Specifically, MMedBench is a multilingual medical multiple-choice QA benchmark across six primary languages: English, Chinese, Japanese, French, Russian, and Spanish. The entire test set of MMedBench comprises 8,518 QA pairs. For a unified evaluation, we remove the samples with multiple answers and use the filtered 8,178 samples as the evaluation set.
From the results in Figure~\ref{fig:mmedbench_all}, we find that \texttt{KDS} brings better performance gains against the other methods across all languages. Specifically, in the Qwen2.5-7B models, compared to the base model, \texttt{KDS} achieves up to \textbf{+4.17\%} average performance gains, especially \textbf{+6.25\%} gains in Russian and \textbf{+3.79\%} gains in Chinese. These results indicate that, by filtering the conflict data, our \texttt{KDS} can better stimulate the LLMs' internal knowledge and boost their multilingual QA performance.

\begin{table*}[t]
\setlength{\tabcolsep}{10pt}
\resizebox{\textwidth}{!}{%
\begin{tabular}{lccccccccccc}
\toprule
Method & CPA & BQ & SQ & FQ & IQ & EA & TP & FQ & CFP & AFM & Avg. \\ \midrule
Base & 40.88 & 53.64 & 49.02 & 56.80 & 51.31 & 61.27 & 36.68 & 51.04 & 50.58 & 32.23 & 48.35 \\
Full-SFT & 38.98 & 53.62 & 49.32 & 55.16 & 52.16 & 61.15 & 35.25 & 49.65 & 50.58 & 32.50 & 47.84 \\
Random & 40.50 & 56.03 & 48.81 & 56.77 & 51.44 & 60.96 & 37.09 & 49.88 & 52.54 & 32.36 & 48.64 \\
IFD & 40.05 & 55.47 & 48.47 & 58.26 & 50.72 & 60.38 & 37.91 & 49.19 & 51.53 & 32.95 & 48.49 \\
Deita & 40.73 & 55.31 & 48.72 & 56.42 & 51.44 & 60.19 & 35.45 & 51.27 & 50.85 & 35.23 & 48.56 \\
3DS & 40.43 & 55.23 & 49.74 & 56.54 & 51.58 & 60.38 & 35.66 & 48.96 & 54.92 & 29.55 & 48.30 \\
\rowcolor{gray!20} \texttt{KDS}-\textit{KA} & 40.35 & 56.83 & 49.23 & 56.65 & 51.01 & 61.54 & 37.50 & 50.12 & 52.20 & 35.23 & 49.07 \\
\rowcolor{gray!20} \texttt{KDS}-\textit{KC} & 39.82 & 55.55 & 49.23 & 57.22 & 49.57 & 60.96 & 37.30 & 49.88 & 52.20 & 35.23 & 48.70 \\
\rowcolor{gray!20} \texttt{KDS}-\textit{KA+KC} & 40.81 & 55.55 & 49.23 & 56.54 & 53.16 & 60.58 & 35.86 & 51.50 & 53.22 & 38.64 & \textbf{49.51} \\
\bottomrule
\end{tabular}
}
\caption{\textbf{Performance comparison (\%) on the Chinese finance-domain QA benchmark, \textit{i.e.}, FinanceIQ}~\cite{zhang2023xuanyuan}. LLaMA3-8B-Instruct is used as the base model in this experiment. ``Avg.'' denotes the average performance.}
\label{tab:finance}
\end{table*}

\subsubsection{Hallucination Alleviation}
According to Pal~\textit{et~al.}~\cite{pal2023med}, despite its effectiveness, instruction-tuning has the side effect of exacerbating the hallucination of LLMs. Here, we investigate this problem by evaluating the tuned LLMs on a popular medical hallucination benchmark, Med-HALT~\cite{pal2023med}. Specifically, MedHalt is a recently-proposed comprehensive evaluation framework designed to evaluate hallucination in medical LLMs. It contains two hallucination tests, \textit{i.e.}, reasoning hallucination tests and memory-based hallucination tests. The former is designed to assess how well an LLM can reason about a given problem by means of False Confidence Test (FCT), None of the Above (Nota) Test, and Fake Questions Test (Fake). The latter focuses on evaluating LLMs' abilities to retrieve accurate information from their encoded training data. In our study, we use the reasoning hallucination tests for evaluation.

Following the original paper~\cite{pal2023med}, we measure the accuracy and pointwise score\footnote{Each correct prediction is awarded +1 point, while each incorrect prediction
incurs a penalty of -0.25 points.} for evaluation. We present the evaluation results of tuned Qwen2.5-7B models in Table~\ref{tab:hallucination}. It can be found that instruction-tuning indeed leads to more serious hallucination, as ``Random'' method causes \textbf{-16.21\%} average score drops. More encouragingly, our \texttt{KDS} can effectively alleviate this side effect and bring up to \textbf{+9.86\%} average score gains against the ``Random'' method. Takeaway: \textbf{\textit{These results prove that our \texttt{KDS} can not only improve the multilingual generalization, but also effectively alleviate the hallucination.}}

\begin{table*}[t]
    \centering
    \begin{tabular}{p{0.98\linewidth}}
        \toprule
\textbf{\textit{Case 1}} \\ \hdashline 
\textbf{Question:} How long can a healthy man expect to live?  \\
\textbf{GPT-4o:} In the United States, the average life expectancy for a healthy man is around {\sethlcolor{light_blue}\hl{76 years}} \ldots \\ \vspace{-1mm}
\textbf{Random:} A healthy man can expect to live for approximately {\sethlcolor{light_red}\hl{78 years}}. \\
\textbf{IFD:} The average life expectancy of a healthy man is {\sethlcolor{light_red}\hl{79 years}}. \\
\textbf{Alpagasus:} The average life expectancy for a healthy man is around {\sethlcolor{light_green}\hl{76 years}} old \ldots  \\
\textbf{DEITA:} The average life expectancy for a healthy man is around {\sethlcolor{light_green}\hl{76 years}} \ldots  \\
\textbf{3DS:} A healthy man can expect to live for approximately {\sethlcolor{light_red}\hl{78.7 years}}, according to the Centers for Disease Control and Prevention (CDC) \ldots  \\ \vspace{-1mm}
\textbf{\texttt{KDS}-\textit{KA}+\textit{KC} (Ours):} The average life expectancy for a healthy man is around {\sethlcolor{light_green}\hl{76 years}} old. However, this number can vary depending on a variety of factors, such as lifestyle choices, genetics, and access to healthcare. It's important to note that life expectancy is not the same as maximum lifespan, which is the longest possible lifespan for a human being. While some people may live into their 90s or even 100s, the average life expectancy is a more realistic estimate of what a person can expect to live to based on current trends and data.  \\
\midrule
\textbf{\textit{Case 2}} \\ \hdashline 
\textbf{Question:} What is the average maximum muscle weight percentage for a 6-foot-tall male that weighs 210 pounds?  \\
\textbf{GPT-4o:} \ldots Muscle mass percentage: For a very muscular person (such as a competitive bodybuilder), muscle mass could make up about {\sethlcolor{light_blue}\hl{40-50\%}} of total body weight
 \ldots \\ \vspace{-1mm}
\textbf{Random:} The average maximum muscle weight percentage for a 6-foot-tall male that weighs 210 pounds is approximately {\sethlcolor{light_red}\hl{40-45\%}}. \\
\textbf{IFD:} For a 6-foot tall male who weighs 210 pounds, the average maximum muscle weight percentage is {\sethlcolor{light_red}\hl{43}}. \\
\textbf{Alpagasus:} The average maximum muscle weight percentage for a 6-foot-tall male that weighs 210 pounds is around {\sethlcolor{light_red}\hl{43-45\%}} \ldots  \\
\textbf{DEITA:} The average maximum muscle weight percentage for a 6-foot-tall male that weighs 210 pounds is approximately {\sethlcolor{light_red}\hl{43\%}} \ldots  \\
\textbf{3DS:} The average maximum muscle weight percentage for a 6-foot-tall male that weighs 210 pounds is approximately {\sethlcolor{light_red}\hl{43\%}} of their total body weight \ldots  \\ \vspace{-1mm}
\textbf{\texttt{KDS}-\textit{KA}+\textit{KC} (Ours):} The average maximum muscle weight percentage for a 6-foot-tall male that weighs 210 pounds is around {\sethlcolor{light_green}\hl{40-50\%}}. This is based on the body composition standards set by the National Health and Nutrition Examination Survey (NHANES) conducted by the Centers for Disease Control and Prevention (CDC). These standards are based on data collected from a large and diverse sample of the US population. The standards provide a range of body composition values for different heights and weights, including the percentage of body fat and the percentage of muscle mass. It is important to note that these standards are based on averages and that individual body composition can vary significantly. \\
\bottomrule
    \end{tabular}
    \caption{
    \textbf{Cases of LLMs' responses on the long-form medical QA benchmark}. For ease of illustration, we simplify the responses of baseline LLMs. Notably, we use the outputs of GPT-4o as the reference. The key information is highlighted, where light blue denotes reference answers, light red denotes wrong responses and light green denotes right responses.
    }
    \label{tab:case_study}
\end{table*}

\subsection{Does \texttt{KDS} can be applied to more domains?}
In the main experiments, we evaluate our \texttt{KDS} in the medical instruction-tuning setting. Here, we investigate whether it can be applied to more domain-specific scenarios, \textit{e.g.}, the widely-interested financial domain.
Specifically, 40K Chinese finance instruction data are collected from the public IndustryInstruction\footnote{https://huggingface.co/datasets/BAAI/IndustryInstruction} dataset as the original training set. We use various DS methods to select 5K data for fine-tuning the LLaMA3-8B-Instruct model. During the implementation of our \texttt{KDS} framework, we use a multilingual NLI model\footnote{https://huggingface.co/MoritzLaurer/mDeBERTa-v3-base-mnli-xnli} to measure the \textit{KA/KC} scores for Chinese instruction data. For evaluation, we use a popular Chinese finance LLM benchmark, \textit{i.e.}, FinanceIQ~\cite{zhang2023xuanyuan}. FinanceIQ is a Chinese evaluation dataset focusing on the financial field, which mainly assesses the knowledge and reasoning abilities of LLMs in financial scenarios. More specifically, FinanceIQ covers 10 major financial categories and 36 minor financial categories, with a total of 7,173 multiple-choice questions. 

Table~\ref{tab:finance} shows the comparative results of LLaMA3-8B models tuned with different methods. We see that our \texttt{KDS} methods still outperform the other DS methods in the finance scenario. More specifically, compared to the ``Full-SFT'' baseline, \texttt{KDS} brings up to 1.67 average performance gains. These results can prove the universality of our \texttt{KDS}.

\subsection{Reliability of NLI models}
\label{sec:appendix_nli}
As aforementioned, we use the DeBERTa-v3~\cite{hedebertav3} model tuned on the MNLI~\cite{williams2018broad} dataset as the NLI models in our \texttt{KDS}. Some readers may wonder whether these NLI models have the ability to identify knowledge alignment and consistency. To investigate this, we manually label 100 pairs of answers and model responses, and evaluate the performance of these NLI models. The results are shown in Table~\ref{tab:nli_manual}, from which we find that larger NLI models achieve better performance, confirming our statements in \S\ref{sec:ablation}. More specifically, the large-size model achieves an accuracy of up to 89\%. Thus, we believe that it is reliable to use them as NLI models in \texttt{KDS}. 
\begin{table}[H]
\centering
\setlength{\tabcolsep}{8pt}
\resizebox{0.48\textwidth}{!}{
\begin{tabular}{lccc}
\toprule
Task & \texttt{xsmall} & \texttt{base} & \texttt{large} \\ \midrule
NLI accuracy & 79\% & 85\% &89\%  \\
\bottomrule
\end{tabular}
}
\caption{\textbf{Performance of NLI models with varied model sizes on the medical-domain test set}.
We manually label 100 pairs of answers and model responses as the test set.}
\label{tab:nli_manual}
\end{table}

Notably, since \texttt{DeBERTa-v3-large-mnli} has achieved remarkable performance and there is a lack of a medical NLI dataset suitable for LLMs, we do not attempt to further fine-tune the NLI model on the medical NLI corpus in this study. Nevertheless, we believe that incorporating more domain-specific knowledge into the NLI models is more effective and has the potential to further boost the effectiveness of our \texttt{KDS}, which is in our future work. 

\subsection{Efficiency of \texttt{KDS}}
\label{sec:appendix_efficiency}
Some readers may worry about the efficiency of our \texttt{KDS} method, as it requires multiple forward passes of the LLM and additional NLI model inference. Actually, compared to other cutting-edge DS methods, the inference overhead introduced by our method is tolerable. For instance, IFD~\cite{li2024quantity} requires twice-forward passes and additional training. DEITA~\cite{liumakes} needs to carefully collect the quality and complexity data and use these data to train additional LLM-based quality and complexity scorers (\textit{e.g.}, LLaMA models). Alpagasus~\cite{chenalpagasus} needs to access a third-party LLM (\textit{e.g.}, ChatGPT), which is expensive and time-consuming. 3DS~\cite{ding20243ds} also requires four forward passes of the LLM. \textit{These powerful counterparts generally sacrifice efficiency for better performance.}

In contrast, although \texttt{KDS} requires two forward-pass processes (\textit{i.e.}, quality filtering and knowledge-aware data scoring) of LLMs and an additional NLI model, we can use some strategies to accelerate the inference. In practice, we can first perform the low-cost quality filtering process and select a relatively small high-quality subset for the subsequent knowledge-aware data scoring. By doing so, the inference budgets can be greatly reduced. Moreover, the NLI checking and diversity filtering processes only require the smaller models, which will not induce much latency. In general, compared to the prior DS methods that rely on heuristic methods (\textit{e.g.}, GPT-4 annotation) or additional training, our \texttt{KDS} is relatively more feasible in real-world applications, and the latency of \texttt{KDS} is tolerable against its performance gains.

\subsection{Case Study}
\label{sec:appendix_case}
In addition to the above quantitative evaluation, we provide some case studies to qualitatively analyze the long-form medical question-answering abilities of LLMs. Specifically, taking two questions in the long-form medical QA benchmark~\cite{hosseini2024benchmark} as examples, we report the comparisons of LLaMA3-8B models tuned with different methods in Table~\ref{tab:case_study}. Specifically, for our \texttt{KDS} method, we use the combined ``\textit{KA}+\textit{KC}'' metric in this study. Since the long-form medical QA benchmark only provides the questions without answers, we present the outputs of a proprietary LLM, \textit{i.e.}, GPT-4o, as reference answers. From the model outputs of Table~\ref{tab:case_study}, it can be found that, LLMs tuned with other DS counterparts could generate inaccurate and too simple responses, as the conflict training data might damage the model's prior knowledge and lead to hallucination. Conversely, with the help of our \texttt{KDS}, LLM can achieve more effective domain adaptation and output more professional and accurate responses. These results continue proving the effectiveness and superiority of our \texttt{KDS} method.

\section{Conclusion}
\label{sec:conclusion}
In this paper, we focus on the knowledge conflict problem in the domain-specific instruction-tuning, which is critical yet under-explored. Specifically, we reveal that fine-tuning LLMs using data contradictory to LLMs' pretrained knowledge would damage LLMs' prior abilities and lead to poor performance. In response to this problem, we propose an innovative knowledge-aware DS (\texttt{KDS}) framework, which involves using two metrics to quantitatively measure the knowledge conflicts from two aspects: i) context-memory knowledge alignment and ii) intra-memory knowledge consistency. By filtering the data with higher knowledge conflicts and sampling the high-quality and diverse data, \texttt{KDS} can effectively stimulate the LLMs' internal abilities and boost the domain-specific performance. Taking the representative medical domain as the testbed, we apply our \texttt{KDS} to select the desired medical instruction-tuning data and use these selected data to fine-tune three widely-used LLMs. Extensive results on several medical question-answering benchmarks demonstrate the effectiveness and universality of our \texttt{KDS}. Moreover, in-depth analyses prove that our \texttt{KDS} method can achieve higher data efficiency and effectively alleviate the model hallucination.

\IEEEpeerreviewmaketitle

\ifCLASSOPTIONcaptionsoff
  \newpage
\fi

\bibliographystyle{IEEEtran}
\bibliography{tkde.bib}

\end{document}